\documentclass{article}
\usepackage[a4paper]{geometry}
\usepackage{amsmath,amssymb}
\usepackage{booktabs}
\usepackage{lmodern}
\usepackage{bookmark}
\usepackage{caption}
\usepackage{subcaption}
\usepackage{appendix}
\usepackage{multicol}
\hypersetup{hidelinks}

\usepackage{authblk}

\usepackage{setspace}
\usepackage{natbib}
\usepackage{graphicx}
\graphicspath{ {./figures/} }
\usepackage{epstopdf}

\title{Classification of geological borehole descriptions using a domain adapted large language model}
\author[1]{Hossein Ghorbanfekr\footnote{Corresponding author: hossein.ghorbanfekr@vito.be}}
\author[1]{Pieter Jan Kerstens}
\author[1]{Katrijn Dirix}
\affil[1]{Flemish Institute for Technological Research (VITO), Boeretang 200, Mol B-2400, Belgium}
\date{\today}

\begin{document}
\maketitle
\begin{abstract}

Geological borehole descriptions contain detailed textual information about the composition of the subsurface. However, their unstructured format presents significant challenges for extracting relevant features into a structured format. This paper introduces GEOBERTje: a domain adapted large language model trained on geological borehole descriptions from Flanders (Belgium) in the Dutch language. This model effectively extracts relevant information from the borehole descriptions and represents it into a numeric vector space. Showcasing just one potential application of GEOBERTje, we finetune a classifier model on a limited number of manually labeled observations. This classifier categorizes borehole descriptions into a main, second and third lithology class. We show that our classifier outperforms both a rule-based approach and GPT-4 of OpenAI. This study exemplifies how domain adapted large language models  enhance the efficiency and accuracy of extracting information from complex, unstructured geological descriptions. This offers new opportunities for geological analysis and modeling using vast amounts of data.

\end{abstract}

\section{Introduction}\label{sec:introduction}
Geological borehole descriptions represent a fundamental datasource in the field of geology.  Collected over many decades, often at great financial expense by geological survey organizations, these borehole descriptions contain detailed textual descriptions of the composition of the subsurface \citep{lawley2023}. They are fundamental for a wide range of applications, from mineral exploration to groundwater management and geotechnical engineering. They are also one of the principal data sources for the construction of geological (3D) models \citep{kaufmann2009}. Although they contain vast amounts of scientific information, they are often stored in poorly accessible (sometimes even analogue) and linguistically unstructured formats. This  means that utilizing these descriptions in models or other applications generally necessitates laborious manual work. Such a requirement substantially hampers utilization of their full potential. The most common way to overcome this is to apply rule-based scripting or manual labelling to transform the unstructured text into lithology classes \citep{hademenos2019, stafleu2016, vanharen2023}.

Taking advantage of recent developments in large language models (LLM), this study develops GEOBERTje: a domain adapted LLM trained on geological borehole descriptions in the Dutch language from Flanders (Belgium). We subsequently finetune a classifier to perform a lithology classification task.

\subsection*{Adoption of LLMs in the field of geosciences}
Emerging advances in the development of large language models have significantly impacted the field of natural language processing (NLP) by enabling machines to generate and comprehend human-like texts. In particular, recent studies have been exploring the potential of NLP to capture geological information from unstructured text. 
Both non contextual models with static word embeddings such as Word2Vec and GloVe are employed \citep{lawleyetal2022, padarian2019} to this aim, as well as more advanced contextual transformer architectures \citep{vaswanietal2017} (e.g., BERT \citep{devlinetal2018}). Transformers have reshaped the NLP domain by introducing an innovative architecture that outperforms traditional methods like recurrent neural networks (RNNs) both in terms of accuracy and scalability~\citep{Vaswani20171y, Howard2018wj}. Unlike RNNs, which process inputs sequentially, transformers employ a self-attention mechanism that allows them to dynamically weigh the importance of different words in a sequence. This mechanism facilitates the generation of contextual embeddings, allowing the model to concentrate on various segments of the input sequence when producing output representations. Consequently, transformers capture long-range dependencies more effectively than RNNs. These advancements have democratized access to pretrained contextual large language models and catalyzed their adoption across diverse domains (including geosciences). 

The main goal of many of the NLP studies within the field of geology is to perform text mining on geological archives and annotate them with meaningful semantic entities. For this purpose, named entity recognition (NER) pathways are frequently developed \citep{rachel_heaven_2020_4181488, enkhsaikhan2021, morgenthaler2022,  qiu2019}. 
In addition, LLM’s have been applied to a range of downstream tasks, such as article summarization \citep{ma2022}, translation \citep{gomes2021}, prospectivity modeling \citep{lawley2023} and mineral exploration \citep{enkhsaikhan2021mineral}. Although English remains the reference language in much of this research, similar applications can be found in other languages such as Chinese \citep{li2021, qiu2018} and Portuguese \citep{consoli2020, gomes2021}.

As far as the classification of borehole descriptions is concerned, the study by \citet{fuentesetal2020} is of specific relevance. In their study, a GloVe model trained on a large corpus of articles in the geosciences domain is used to obtain embeddings of drill core descriptions. Next, a multilayer perceptron neural network is trained to classify them. Finally, they created 3D maps by interpolating the embeddings. \citet{fuentesetal2020} utilize a non-contextual model with static word embeddings.

\subsection*{The advantages of domain adaptation}
Research has shown that LLMs like BERT, primarily trained on general English language corpora, face inherent limitations when applied to domain-specific classification tasks involving texts in other languages. The primary reason is that these models have not been exposed to sufficient examples of technical language in the target language. This limits their effectiveness to recognize and process such texts. Monolingual models such as Camembert~\citep{martin2019} for French and BERTje~\citep{devriesetal2019} for Dutch outperform the multilingual BERT model on downstream NLP tasks \citep{devriesetal2019, martin2019}.
More recently, ChatGPT represents a significant advancement in the field of natural language processing with its advanced generative capabilities and ease of accessibility. Its design allows for flexible interaction through prompt engineering: a method that requires considerably less technical expertise compared to the complexities involved in finetuning models like BERT. This accessibility makes ChatGPT an attractive option for scientists who may not specialize in machine learning or who do not have access to specialized GPUs. This provides a more user-friendly approach to harness the power of large language models for domain-specific tasks \citep{openai2024gpt4}.

\subsection*{Scope}
In this study we develop GEOBERTje: a domain adapted version of BERTje \citep{devriesetal2019} (a pretrained Dutch BERT model). The domain adaption is achieved by transfer learning on a large corpus of $283 000$ unlabeled Flemish borehole descriptions. Subsequently,  we use a much smaller set of approximately $2 500$ labeled samples to fine tune this model to a lithology classifier that is able to extract multiple lithology classes from borehole descriptions. We compare the performance of the classifier based on GEOBERTje with those obtained using (a) traditional rule-based scripting and (b) GPT-4 of OpenAI through prompt engineering. Our study exemplifies how the adoption of domain-specific LLMs can provide significant added value in the field of geology by enhancing the efficiency and accuracy with which unstructured geological datasets can be analyzed.

\section{Data and setting}\label{sec:data}
The importance of geological data, combined with insights of expert-geologists, to develop robust and accurate NLP models for interpreting lithological descriptions cannot be overstated. High-quality, interpreted, geological data provides the foundational knowledge and context necessary for training and finetuning machine learning models, ensuring they can effectively interpret and differentiate between various lithological features. 
In the following sections we discuss the study area, the available data sources in Flanders, and the preprocessing steps undertaken to prepare the training, validation and test set for this study.

\subsection{Borehole descriptions in Flanders}\label{dov}
The majority of Flemish borehole descriptions reside in Databank Ondergrond Vlaanderen (DOV\footnote{\url{https://www.dov.vlaanderen.be}}), which is a partnership among various government entities concerned with Flanders' subsurface. Initiated in 1996, it is now the main data holder for open Flemish data and information concerning geology, natural resources, soil, hydrogeology, geotechnical characteristics and groundwater licenses. It contains digitized archives of several federal, regional and research institutes and is constantly updated with new data~\citep{denil2020}. DOV offers unparalleled access to Flanders' subsurface information with a relational database backbone, a web service for interactive querying and visualization and an API that supports machine-based extraction \citep{denil2016, haest2018}. In addition to serving as an information platform, it also acts as a prime data source for regional and cross-border 3D geological models \citep{deckers2019, vernes2018}. A model type specifically in need of large volumes of subsurface data are the geological \textit{voxel} models, composed of volumetric pixels that contain modeled data on the load of a specific lithology (peat, clay, gravel, etc.)~\citep{vanharen2023}. At the base of these voxel models lie tens of thousands detailed standardized borehole descriptions. While the borehole descriptions are included in a structured format with the DOV database\footnote{For example, see \url{https://www.dov.vlaanderen.be/data/interpretatie/2016-252085}}, the description intervals themselves are generally composed of unstructured text. In-house rule-based scripts were developed to classify them into detailed categories to enable direct usage of these descriptions into models. These rule-based scripts employ dictionaries and regular expressions to transform the descriptions into a primary, secondary and tertiary lithology class, as is displayed in Table~\ref{tab:classification} \citep{vanharen2016, vanharen2023}. While admixtures are generally captured as well, they fall outside the scope of this paper, as they are easily extractible using traditional rule-based text mining methods.

\begin{table}[ht]
\centering
\resizebox{0.9\columnwidth}{!}{
\begin{tabular}{|p{6cm}cccc|}
\hline
\textbf{Original description} & \textbf{Main lithology} & \textbf{Secondary lithology} & \textbf{Tertiairy lithology } & \textbf{Admixture} \\[0.5em] 
\hline
Greenish yellow loam with sand (mostly coarse) and gravel & Loam & Coarse Sand & Gravel & \\[0.5em] 
\hline
Fine sand with a very small fraction of gravel, and abundant intercalations of clay & Fine Sand & Clay & Gravel & \\[0.5em] 
\hline
Reddish-brown sand, very poorly sorted, very fine, with a lot of admixture of coarse sand (grains of $200$-$300$\,\textmu m even up to $1$\,mm), slightly coarser at the bottom. A rare weathered shell. & Fine Sand & Coarse Sand & & Shell \\[0.5em] 
\hline
\end{tabular}
}
\caption{Examples of unstructured lithological descriptions transformed into structured categories. The original descriptions were translated to English for reasons of clarity.}
\label{tab:classification}
\end{table}

\subsection{Flanders' shallow geology}
The geology of Flanders is characterized by a relatively flat landscape dominated by Quaternary deposits of sands, gravels, and clays, along with localized peat bogs. In the southern part of Flanders, a Pleistocene loess belt is present \citep{vanharen2016}. These deposits overlie 
Neogene or Paleogene sediments, which are rich in sand and clay layers, deposited in various marine and continental Neogene and Paleogene environments. Beneath these, Mesozoic and Paleozoic deposits feature a complex assembly of older marine, peri-marine and continental rock formations, including limestone and shale. The shallow subsurface is hence almost entirely composed of soft sediments, with a predominance of sandy and clayey deposits. The lithology classes applied to categorise these soft sediments are adapted from \citet{wentworth1922} and given in Table~\ref{tab:lithoclass}. In this table different grain size classes of sand are distinguished, as each of these classes have different geotechnical characteristics and contrasting applications as raw materials \citep{flemish2010}. In addition, a class \textit{sand} without further grain size specification is included. Although the presence of this class poses challenges for subsequent lithological modelling tasks, it has to be present as many of the lithological descriptions do not mention grain size information. For example, \textit{green sand with clay and some gravel at the base}. 
\begin{table}[h!]
\centering
\begin{tabular}{|l|l|}
\hline
\textbf{Lithoclass (Dutch)} & \textbf{Lithoclass (English)} \\ \hline
Veen                         & Peat                          \\ \hline
Klei                         & Clay                          \\ \hline
Silt                         & Silt                          \\ \hline
Leem                         & Loam                          \\ \hline
Fijn zand                    & Fine sand                     \\ \hline
Middelmatig zand             & Medium sand                  \\ \hline
Grof zand                    & Coarse sand                   \\ \hline
Zand                         & Sand                          \\ \hline
Grind                        & Gravel                        \\ \hline
\end{tabular}
\caption{Lithology classes relevant for the Flemish shallow subsurface.}
\label{tab:lithoclass}
\end{table}

The severe imbalance in the occurrence of these different lithology classes in Flanders can be visualized by inspecting the distribution of each lithology class for the main lithology or secondary lithologies as determined by the rule-based scripting. 
Figure~\ref{fig:classcount_unlabeled} shows that clay (\textit{klei}), and the different sand fractions (\textit{zand},  \textit{fijn zand}, \textit{middelmatig zand} and \textit{grof zand}) are the predominant main lithologies.  It further reveals that secondary lithologies are less commonly documented, with tertiary lithologies being even rarer. This is logical given the fact that only the most detailed descriptions document up to three lithologies within a single lithological interval. Nonetheless, capturing these different levels of detail can be essential to fully understand  the complexity of the shallow subsurface, as the detailed descriptions in Table~\ref{tab:classification} illustrate.

\subsection{Data preparation}\label{preprocessing}
\begin{figure}[htbp]
	\centering
	\includegraphics[width=0.8\textwidth]{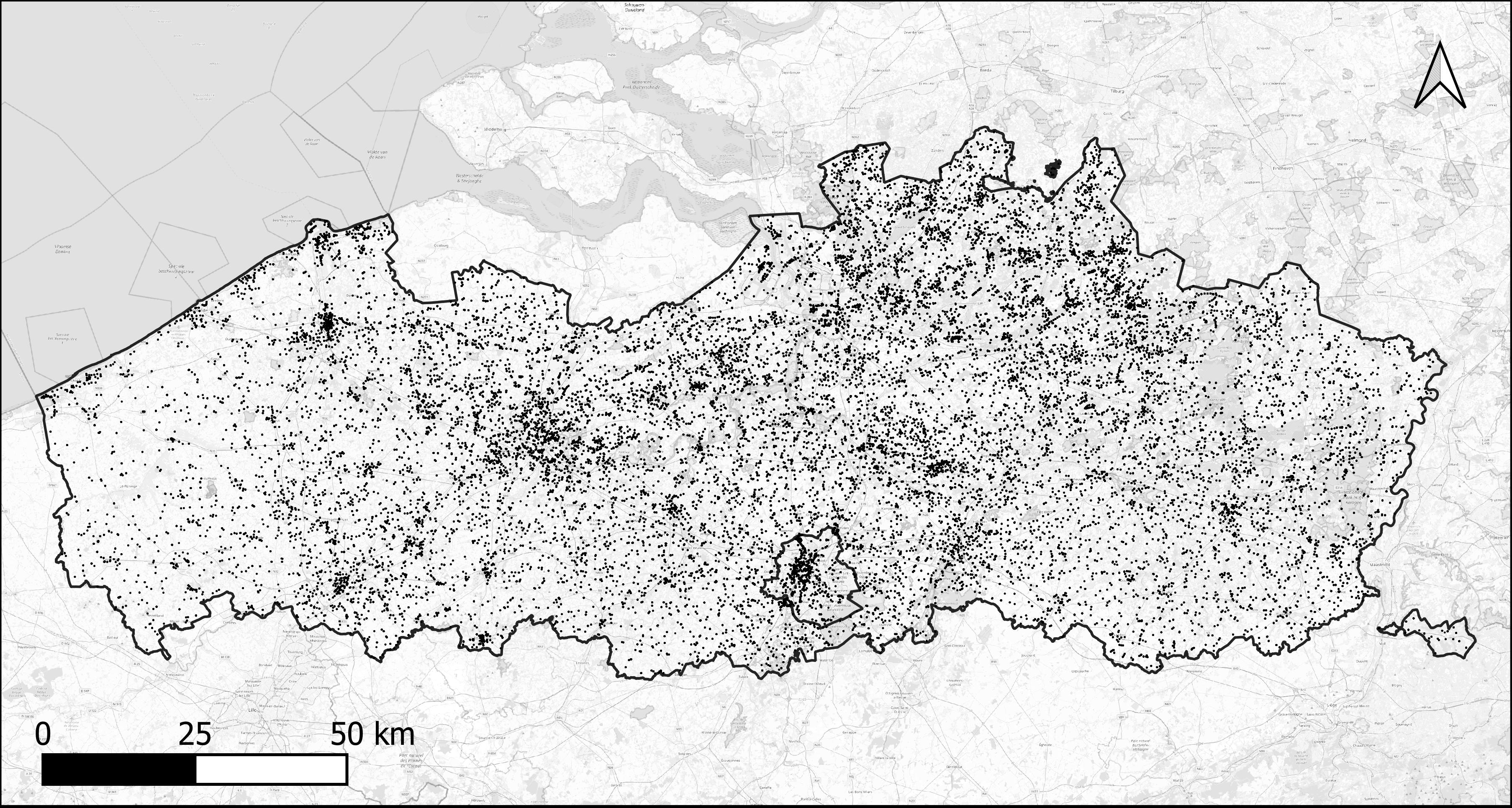}
	\caption{Spatial distribution of the sampled borehole descriptions in Flanders and Brussels.}
	\label{fig:spatial_distribution}
\end{figure}

We selected a total of $341 000$\, borehole description intervals originating from $23 000$\, boreholes spatially spanning the entire territory of Flanders. Figure~\ref{fig:spatial_distribution} shows the spatial distribution of the sampled borehole descriptions in Flanders and Brussels. Although boreholes are spread throughout Flanders, a larger concentration can be observed in more densely populated areas such as Brussels, Antwerp, Ghent, etc. This initial dataset contains descriptions in both Dutch and French.

The following preprocessing steps were applied 
to improve the quality of the dataset: (i) removing all French descriptions; (ii) removing descriptions pertaining to multiple depths of the borehole; (iii) replacing \textit{idem} with the borehole description immediately preceding it.
The final dataset contains $283 000$\, borehole descriptions. The average length of the lithological descriptions is $19$ words with a standard deviation of $26$. The minimum and maximum number of words per description is $1$ and $336$. 

\begin{figure}[htbp]
	\centering
	\begin{subfigure}[b]{0.45\textwidth}
		\includegraphics[width=\textwidth]{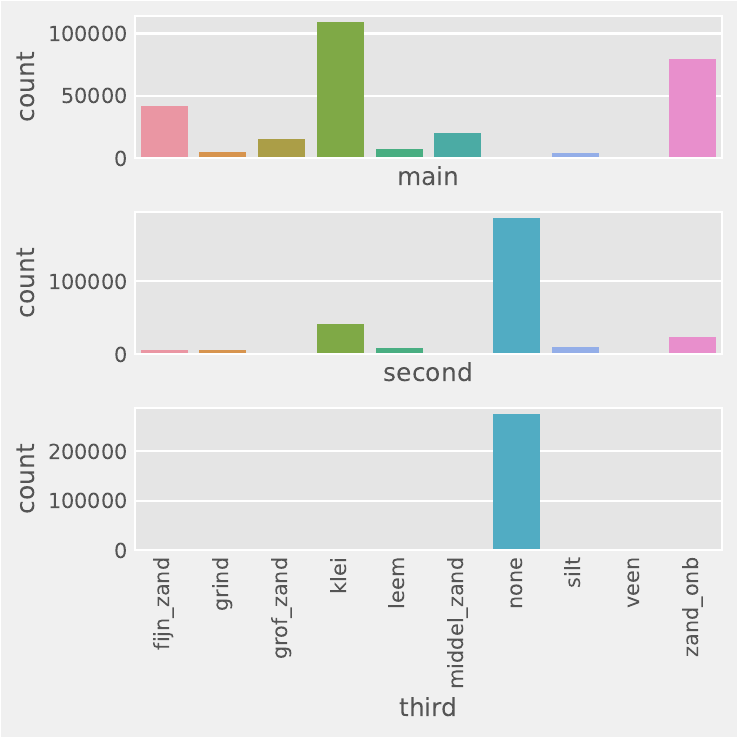}
		\caption{Full dataset.}
		\label{fig:classcount_unlabeled}	
	\end{subfigure}
	\begin{subfigure}[b]{0.45\textwidth}
		\includegraphics[width=\textwidth]{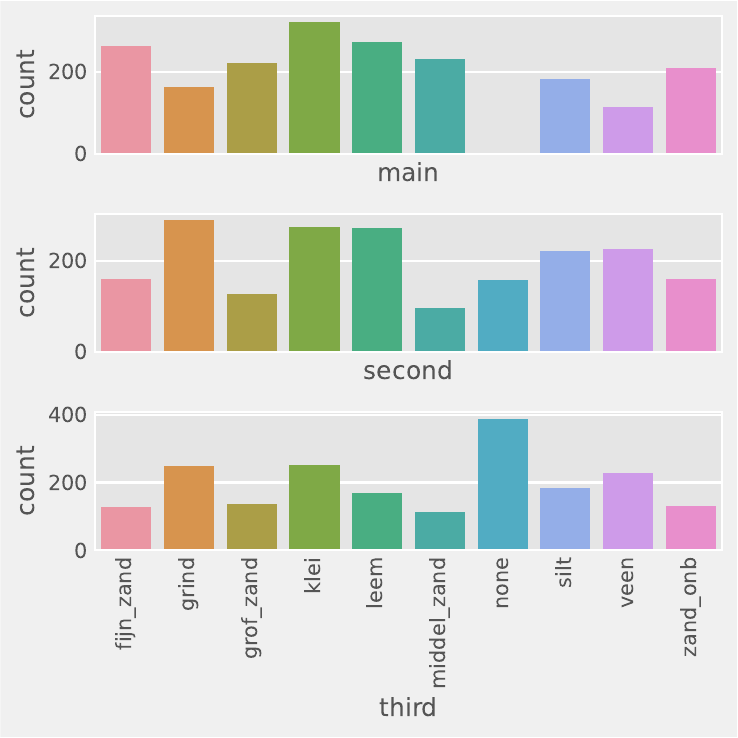}
		\caption{Selected subset for ground truth labeling.}
		\label{fig:classcount_labeled}
	\end{subfigure}
	\caption{Occurrence of lithology classes according to the rule-based script.}
\end{figure}

Labeled data are required to develop the lithology classifier. 
Rule-based scripts are available to directly classify this dataset into a main and secondary lithologies (\textit{i.e.,} second and third)~\citep{vanharen2023}. However, directly using these labels in training of a classification model would at best only hope to achieve a performance on par to these rule-based scripts. Therefore, a ground truth needs to be established through manual labeling by experts. In order to make most efficient use of their time, a limited subset needed to be selected for manual labeling. However, random sampling from a dataset with such an imbalanced distribution  would replicate this bias leading to classifiers that can only perform well on the majority classes because they have not seen enough training observations of the minority classes (Figure~\ref{fig:classcount_unlabeled}).
Therefore, we first eliminated duplicate observations and then grouped the remaining data by (rule-based) lithology class to select descriptions for manual labeling. Subsequently, we randomly chose a similar number of samples from each of these groups. This resulted in $2671$ borehole descriptions that were manually labeled. Finally, we partitioned this dataset into
three subsets: $15$\% for testing, $15$\% for validation, and the remaining $70$\% for training.

Figure~\ref{fig:classcount_labeled} shows the lithology class occurrence (as labeled by the rule-based script) of the selected subset. The class distribution is still skewed but far less so than before. This subset was subsequently labeled manually by experts.

\section{Methods}\label{sec:methods}
\begin{figure}[htbp]
	\centering
	\includegraphics[width=0.9\textwidth]{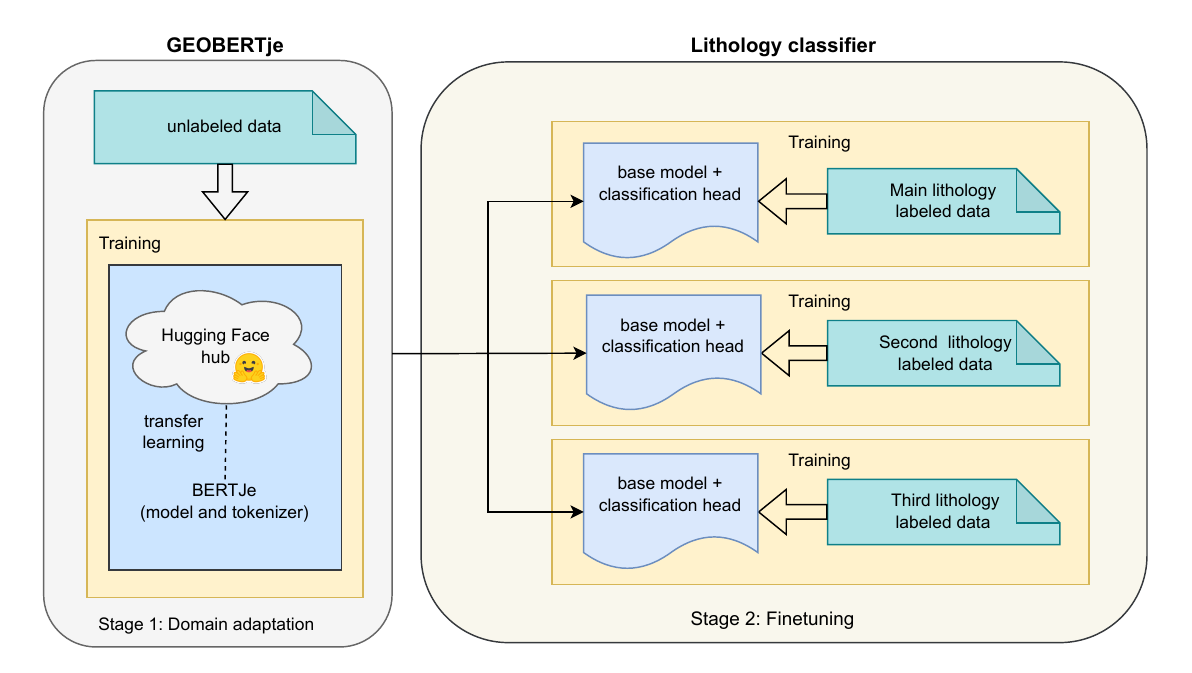}
	\caption{Diagram depicting the two-stage training workflow of GEOBERTje for the lithology classification task, 
	utilizing both unlabeled (stage 1) and labeled data (stage 2).} 
	\label{fig:training-workflow}
\end{figure}

We utilize a transformer architecture to create a domain adapted LLM on Dutch geological borehole descriptions in Flanders (GEOBERTje) which we further train to perform a lithology classification task. We use the Hugging Face platform to acquire the base model and carried out training in two main stages~\citep{tunstall2022natural,wolf-etal-2020-transformers}. Figure~\ref{fig:training-workflow} shows a diagram of the entire training workflow for the lithology classification.

Model training involves two main stages to achieve optimal classification accuracy:
(1) we use BERTje as a base model, tailoring it to the geological domain by 
enhancing its understanding of Dutch geological terminology. We call the result GEOBERTje.
(2) We finetune a classifier head based on GEOBERTje, 
utilizing the assigned labels to refine its ability to accurately classify different lithologies.

\subsection{Training stage 1: Domain adaptation}
We start from BERTje of \citet{devriesetal2019}.\footnote{The model is available \href{https://huggingface.co/GroNLP/bert-base-dutch-cased}{here} on the Hugging Face platform.} BERTje uses the same model architecture as the original BERT model of \citet{devlinetal2018}. BERT is trained on a multilingual text corpus. While only Wikipedia articles in Dutch belong to the Dutch training corpus of multilingual BERT, BERTje was trained on a large and more diverse corpus of 2.4 billion Dutch tokens originating from Dutch books, news, etc \citep{devriesetal2019}. Intuitively, BERTje has an improved performance for the Dutch language compared to multilingual BERT.

We could directly train BERTje with labeled data to perform the classification task. However, since  BERTje's training data is not tailored specifically to geology, refining its understanding of geological subtleties can be achieved through additional training on an unlabeled dataset focused on lithological descriptions.
This method is known as \textit{domain adaptation} (DA). Instead of training a LLM from scratch, it allows further training on data from a particular domain \citep{Guo2022ta}. This pre-training step enhances the model's performance by improving the vector embeddings. This aligns the embeddings more with the downstream classification task (see subsection~\ref{subsec:resultsda}).

DA techniques prove particularly advantageous in scenarios where there is a scarcity of labeled data for a target task but unlabeled data is abundantly present. Training in this approach relies on the masked language model (MLM) objective and does not require labeled data. It involves randomly masking words in the descriptions and predicting the masked word using the model. 
The loss function value is then calculated using the cross-entropy between the labels and logits. Consequently, the model becomes adept at predicting masked words from their context and recognizing underlying patterns and long-range dependencies within the lithological descriptions. This step effectively leverages the vast amounts of unlabeled lithological descriptions available, enriching the model's training process \citep{devriesetal2019}.
Table~\ref{tab:random_masking} illustrates the process of applying random masking to 
an example sentence, ``\textit{weinig fijn zand met grindelementen}''
\footnote{slightly fine sand with shell gravel elements.}.  
The tokens resulting from this process are subsequently utilized in masked language modeling. 
It is important to note that tokens with a label value of $-100$ are 
disregarded by the loss function. Thus, they do not contribute to the training process and 
ensure that the focus remains solely on the missing words in the sentence. 
\begin{table}
\centering
\resizebox{0.9\columnwidth}{!}{
\begin{tabular}{lllllllll}
\toprule
 & 1 & 2 & 3 & 4 & 5 & 6 & 7 & 8 \\
\midrule
\textbf{Original tokens} & [CLS] & weinig & fijn & zand & met & grind & \#\#elementen & [SEP] \\
\textbf{Masked tokens} & [CLS] & weinig & fijn & [MASK] & met & [MASK] & \#\#elementen & [SEP] \\
\textbf{Labels} & -100 & -100 & -100 & 22664 & -100 & 12893 & -100 & -100 \\
\bottomrule
\end{tabular}
}
\caption{An illustration of ``\textit{weinig fijn zand met grindelementen}'' tokenization used in masked language modeling (MLM).} 
\label{tab:random_masking}
\end{table}
We employed a masking probability of $0.15$, which was randomly applied to input sample 
data on-the-fly, with training batch sizes of $32$ and a learning rate of $0.0005$.
\begin{figure}[htbp]
	\centering
	\includegraphics[width=0.6\textwidth]{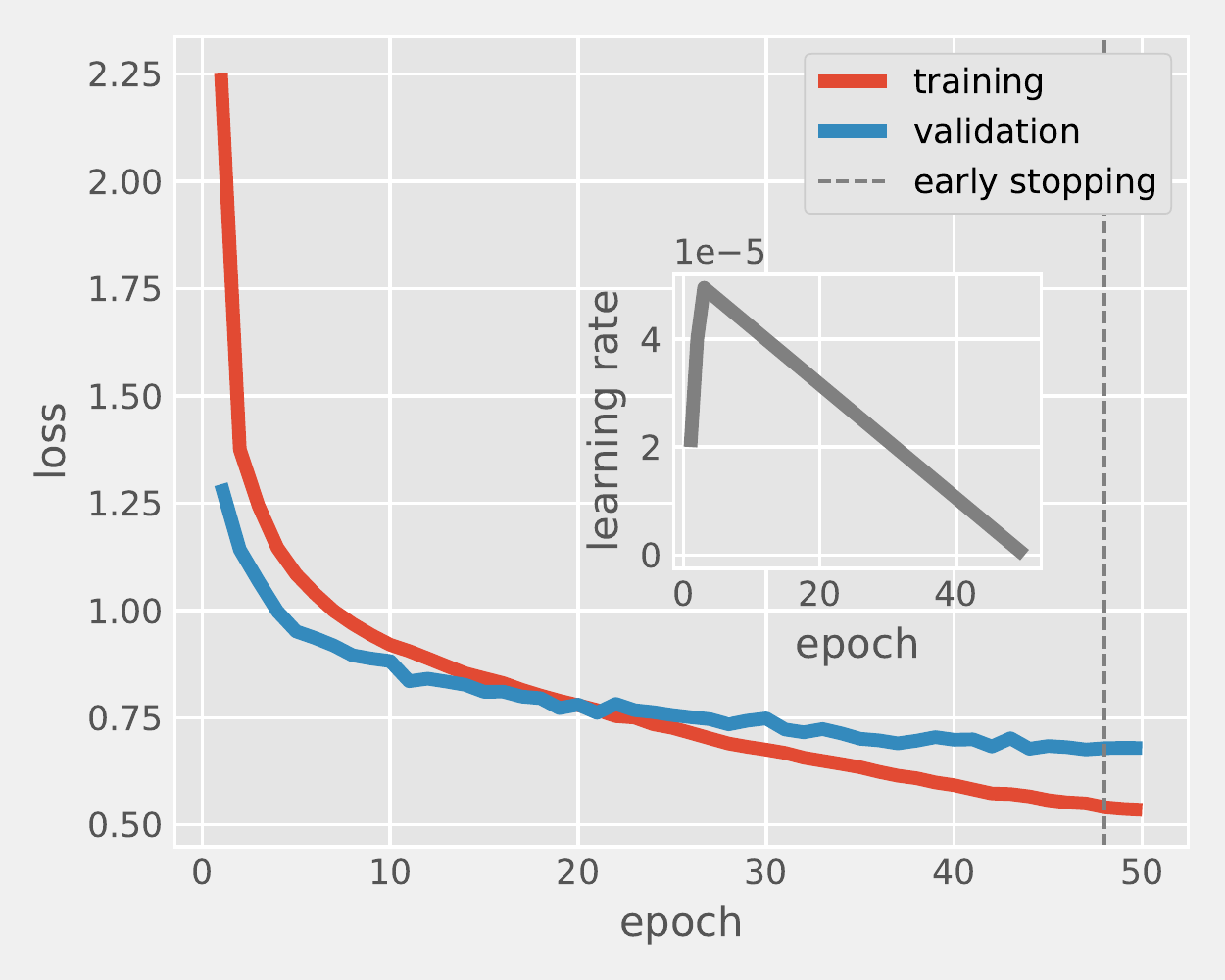}
	\caption{GEOBERTje domain adaptation training (red) and validation (blue) loss curves over epochs. Inset figure: learning rate decay as a function of epoch.}
	\label{fig:pretraining}
\end{figure}

Figure~\ref{fig:pretraining} illustrates how the loss values for both the training and validation sets 
evolve in relation to the number of epochs (i.e., training iterations) during the MLM training.
Initially as the model learns from the data, there is a notable decline 
in loss values: this indicates rapid improvement in predicting masked tokens within the training data. 
This decline stabilizes as the training progresses: this reflects the model's convergence towards optimal performance.
The loss values for the validation set exhibits a similar trend, albeit with occasional fluctuations.
It must be noted that while training loss reflects the model's fit to the training data, 
validation loss assesses its ability to generalize to unseen data. 
If validation loss stagnates or starts to increase, indicating potential overfitting, 
then early stopping mechanisms are employed. 
This involves halting training if there is no improvement in validation loss over a predefined number of epochs. This
prevents the model from overlearning the training data and ensuring better generalization performance.
The inset in Figure~\ref{fig:pretraining} displays the progression of the learning rate during training.
Initially, we implemented a warm-up learning rate to ensure the model's stability and mitigating the potential 
loss of previously acquired knowledge from the BERTje base model.
Subsequently, we applied learning rate annealing to enhance the model's final performance.
Following this step, we possess a pre-trained base language model for Dutch texts that is already 
proficient in lithological descriptions. This serves as a foundation model for the next
finetuning stage and is particularly tailored to the lithology classifications task using labeled data.
\subsection{Training stage 2: Finetuning}
After the initial training phase of domain adaptation, the subsequent step involved finetuning the domain adapted
model using a limited set of labeled data \citep{Howard2018wj}. 
For the classification task, each lithological description accordingly needs to be assigned three labels: a main lithology and up to two secondary lithologies.
``None'' is used if no label was present. 
We chose a simple approach by training three separate classification models for each of the three label groups (illustrated by the yellow parallel boxes in ``Stage 2: Finetuning'' of  Figure~\ref{fig:training-workflow}). 
This means that, starting from the same pretrained base model, all three classification models operate independently (and parallel) from each other.
For each classification model a classification head is attached to the
pretrained base model by adding a linear layer on top. The three models
are further finetuned on the labeled training data to minimize the
cross-entropy loss.
In order to account for potential class imbalances in the labeled data, we utilized class weights
separately for each of the three target labels (see Equation~\ref{eq:weighted-cross-entropy-loss}).
Because class weights assign higher importance to underrepresented classes during training, they reduce
the impact of class imbalances and ensure that the model learns to generalize evenly well across all classes.

\begin{figure}[htbp]
	\centering
	\includegraphics[width=1.0\textwidth]{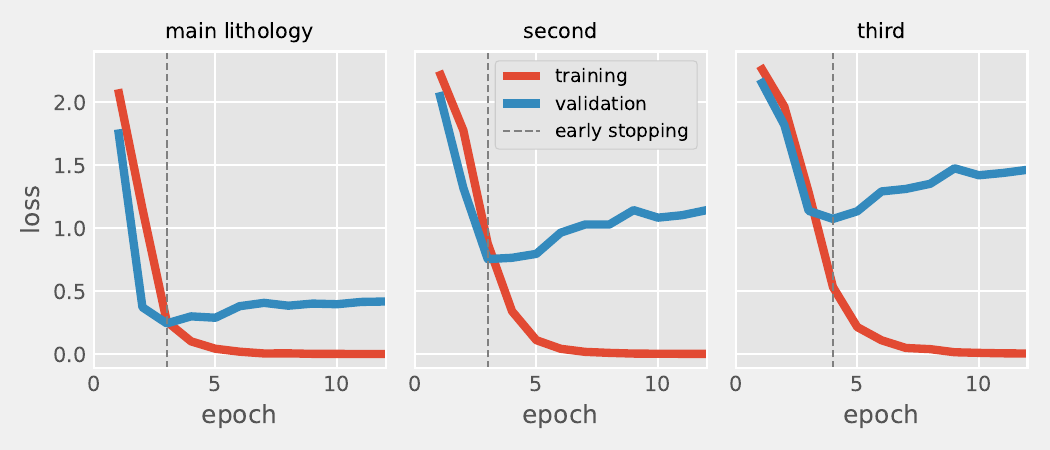}
	\caption{GEOBERTje model fine tuning training (red) and validation (blue) loss functions for the main, second and third lithology class.}
	\label{fig:finetuning}
\end{figure}

The cross-entropy loss function with class weights is defined as:
\begin{equation}\label{eq:weighted-cross-entropy-loss}
\text{loss}(y, \hat{y}) = - \frac{1}{N} \sum_{i=1}^{N} \sum_{j=1}^{C} w_{j} \times y_{ij} \times \log(\hat{y}_{ij}),
\end{equation}
where $N$ spans minibatch dimension, $C$ is the number of classes, 
$y_{ij} = 1$ if the sample $i$ belongs to class $j$, otherwise $y_{ij} = 0$.
$\hat{y}_{ij}$ is the softmax predicted probability of sample $i$ belonging to class $j$.
$w_{j}$ is the weight assigned to class $j$ \citep{Bengio2016-nu}.
In our experience, using class weights yields more accurate results compared 
to oversampling or undersampling methods to deal with uneven data distributions. 
Figure~\ref{fig:finetuning} sets out the cross-entropy loss on the training and validation 
set as a function of the number of epochs  for the main, second and third lithology model. 
While the training loss exhibits an ever decreasing trend with the number of epochs, 
the validation loss first decreases up to an inflection point before increasing again. 
This illustration of the bias-variance trade-off shows that the models learn from the data 
until at a certain point they start overfitting on the training data and their generalization 
ability starts to diminish.
Therefore, training is stopped as soon as the loss on the validation set starts to increase. 
It must be noted that we reused the same learning rate strategy throughout, 
incorporating both warm-up and linear annealing techniques, with a maximum learning rate value set at $0.0005$.
This process results in the final classification models tailored to the lithology classification task. 

\subsection{Postprocessing}\label{postprocessing}
Two postprocessing steps were introduced to further improve the classifiers' overall accuracy and to correct for the fact that the three parallel classifiers are not aware of each other's prediction.
First, a lithology class can occur only once in each classification result. If the predicted secondary (tertiary) lithology equals the predicted main (secondary) lithology, then these duplicate class predictions from the secondary lithologies are removed and replaced by the subsequent most probable class within the group.
Additionally, only model predictions surpassing a confidence threshold of $\tau = 0.1$ are considered valid. This threshold $\tau$ is determined so that it maximizes the accuracy score on the validation set, averaged over each classifier.\footnote{This value did not change when maximizing a different metric such as balanced accuracy or matthews correlation coefficient.} If no classes meet this criterion, then ``None'' is predicted with confidence score $-1$.

\subsection{Other methods: rule-based scripts and GPT-4}
We compare the performance of our domain adapted and finetuned GEOBERTje classifier to classification using the original rule-based script and GPT-4 through prompt engineering. Appendix~\ref{othermethods} describes both approaches in more detail.

\section{Results and discussion}\label{sec:results}

\subsection{Impact of domain adaptation}\label{subsec:resultsda}
\begin{figure}[htbp]
	\centering
	\includegraphics[width=0.6\textwidth]{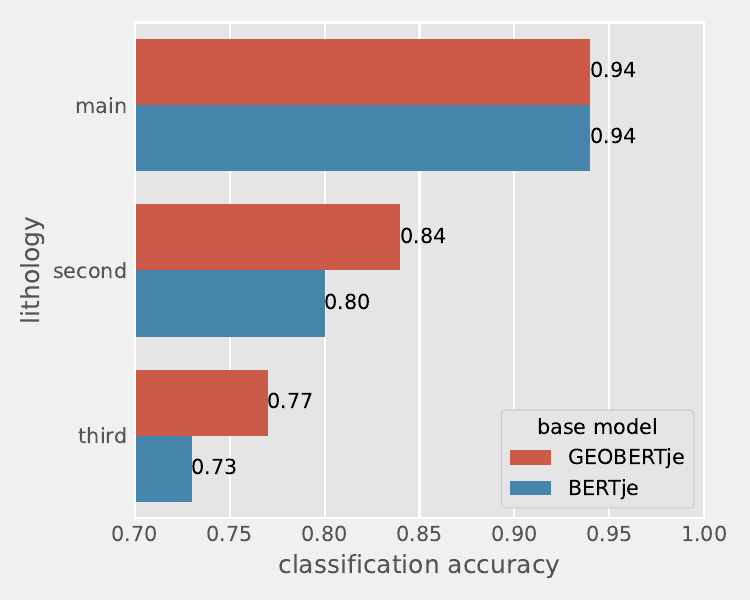}
	\caption{Classification accuracy using two different base models: GEOBERTje (red) and BERTje (blue).}
	\label{fig:base_model}
\end{figure}

We finetuned the classification model directly from the generic BERTje base model (i.e., without domain adaptation) to asses the effect of domain adaptation on accuracy. We subsequently compare the performance of this approach to GEOBERTje. 
Figure~\ref{fig:base_model} compares the classification accuracy of the final finetuned model 
using two different base models: GEOBERTje and the original BERTje. 
The results are presented separately for each label. 
This figure shows a negligible difference in the accuracy score for the main lithology. This suggests that geological context may not be crucial for predicting the main lithology.
However, the domain adapted model demonstrates an improved accuracy for the secondary ($0.84$ vs.~$0.8$ with BERTje as base model) and tertiary lithology ($0.77$ vs.~$0.73$ with BERTje as base model).
This highlights the benefits of employing a domain adapted model over a generic one 
in order to achieve better performance in downstream applications. 
The observed improvement is attributed to the domain adaptation process which enhances the model's understanding of the geological context by taking advantage of unlabeled data. 
This adaptation is especially beneficial to more accurately classify secondary and tertiary lithologies.

\subsection{Classification performance}
\begin{figure}[htbp]
	\centering
	\includegraphics[width=0.6\textwidth]{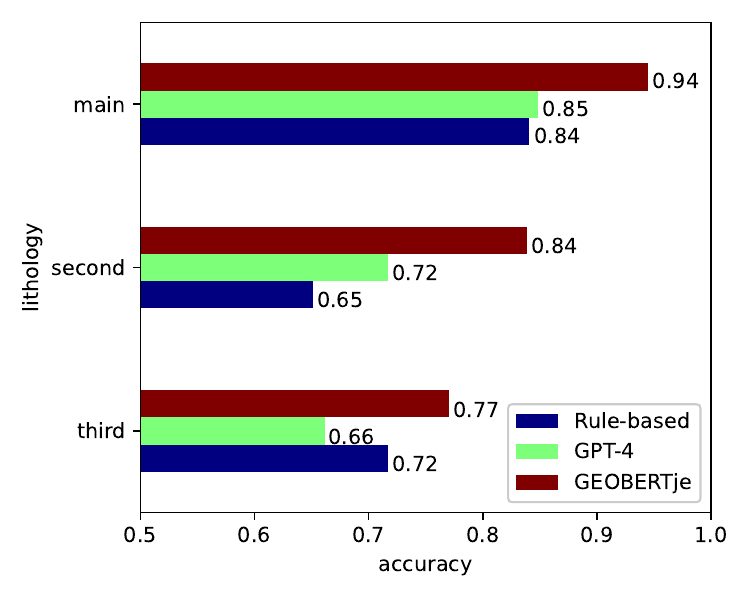}
	\caption{Comparative accuracy of rule-based, GPT-4, and GEOBERTje models in classifying main, secondary, and tertiary lithology from geological drill core descriptions.}
	\label{fig:accuracies}
\end{figure}

\begin{figure}[htbp]
	\centering
	\includegraphics[width=1.0\textwidth]{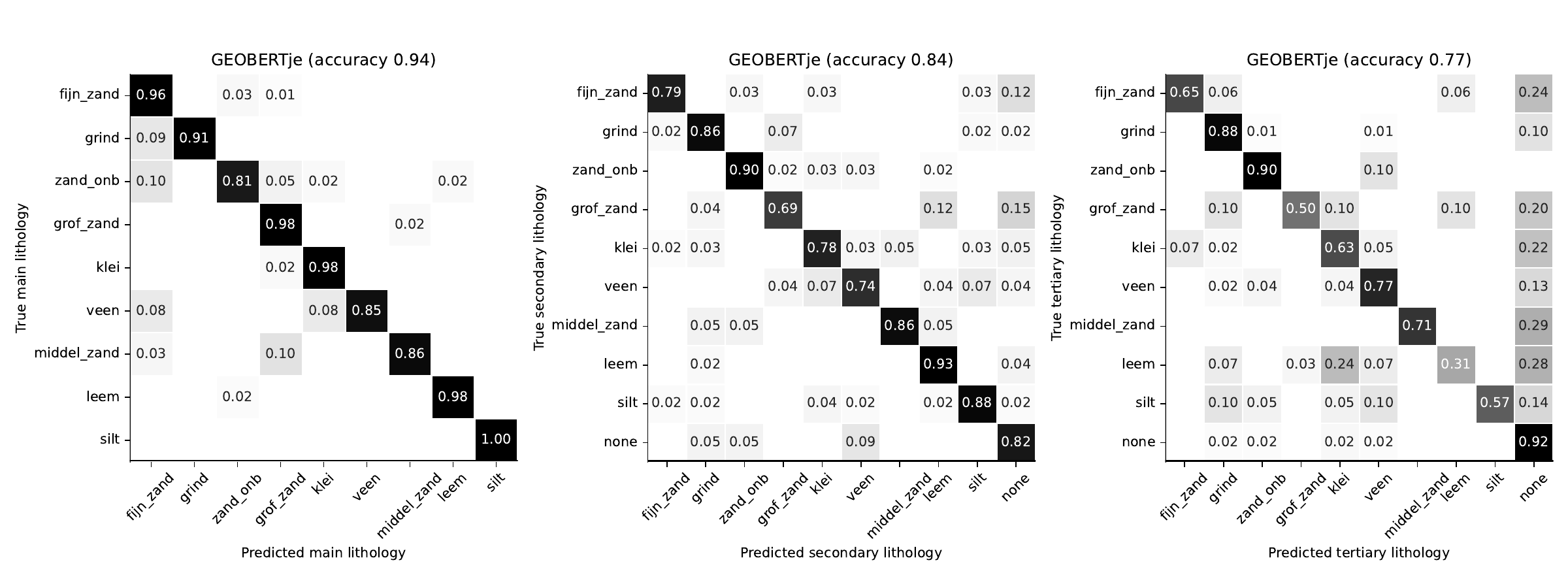}
	\caption{Confusion matrices.}
	 \label{fig:confusion_matrices_geobertje}
\end{figure}

We evaluate the efficacy of our finetuned GEOBERTje classifier against the traditional rule-based method and GPT-4 on lithology classification. First, we consider the classification accuracy -- defined as the ratio of correctly classified observations over the total number of observations -- on an unseen test set  for the main and secondary lithologies. Figure~\ref{fig:accuracies} demonstrates that the GEOBERTje classifier outperforms the other models in all categories. It achieves a very high classification accuracy of $0.94$ on the main lithology.
It exhibits much better performance over the rule-based and GPT-4 models by margins of $9$, $12$ and $6\%$ compared to the runner-up for the main, secondary, and tertiary lithologies respectively.
While GPT-4 competes closely with the rule-based approach in main lithology classification and exceeds it in classifying the secondary lithology, it falls behind in accurately classifying tertiary lithology.
A noticeable pattern is the diminution of accuracy across all models from main to tertiary lithology.
The confusion matrix in Figure~\ref{fig:confusion_matrices_geobertje} facilitates a more detailed analysis of the classification performance for the individual lithology classes. Diagonal entries reveal normalized classification accuracy on a class level, off-diagonal values provide insights on the classes with whom the classifier ``confuses'' the correct label. Figure~\ref{fig:confusion_matrices} in Appendix \ref{appendix: confusion matrix} also contains the confusion matrices for the rule-based scripting and the GPT-4 model.

\subsubsection*{Discussion}
A detailed analysis of the patterns observed in Figures \ref{fig:accuracies} and \ref{fig:confusion_matrices_geobertje} allows to make some general observations. We use concrete examples from the borehole descriptions to better illustrate the nuanced complexities that explain certain patterns.

Both GEOBERTje and GPT-4 are able to correctly interpret linguistically complex phrases where rule-based scripts fail. This occurs, for example, where the adverb denoting the sand fraction is separated from the noun by several words, or occurs after the noun (e.g. \textit{``sand, very fine''}). As an example, the phrase  \textit{``light brown, not calcareous, very strongly peaty, few mica containing, medium fine loamy sand, with semi-deteriorated plant remains''} is correctly classified as  \textit{``fine sand'', ``loam'', ``peat''} for the main, second and third lithology class respectively. This results in higher overall accuracy scores and a higher performance on the sand-classes in particular.

The lower performance of the second and third lithology class for each of the models is apparent. The main explanation for this observation is threefold: 
\begin{enumerate}
\item Correctly interpreting these classes is more complex than capturing the main lithology: the more detailed descriptions often mark down more than three different lithologies.
For example, the phrase \textit{``very coarse, glauconite-rich sand and angular, fine gravel, in a matrix of medium to fine sand''} is challenging to label, even manually. Nonetheless,  GEOBERTje generally outperforms the two other models in correctly handling these complex cases. 
In descriptions where lithologies are described using the word ``to'' (e.g.  \textit{``silty fine to medium sand, with a lot of gravel and some...''}), the rule-based model and GPT-4 tend to pick only one of the two (sand) classes, while GEOBERTje more often captures them both.
\item The dataset is skewed in terms of lithology class occurrence, in particular for the second and third lithology class (see subsection \ref{preprocessing}). In addition, both the test- and training set are limited in size (approx. $400$ and $1870$ samples, respectively). Consequently, some labels are only scarcely represented in the training and test set of the second and third lithology term. As a result, one of the most common errors of the classifiers of these groups is to miss certain labels and predict the much more frequently occurring ``None'' class. This also leads to less reliable normalized accuracy scores in the detailed confusion matrices (Figure~\ref{fig:confusion_matrices}).
\item The choice of assigning a label to the secondary or the tertiary lithology class is challenging and can be ambiguous as they are sometimes interchangeable (e.g., \textit{``sand with very little gravel and some silt''}). The rule-based scripts particularly struggle to correctly make the distinction between these two groups, resulting in lower accuracy scores.
\end{enumerate} 

Despite our best efforts to correctly label this data, some errors persisted. This is an issue that is nearly unavoidable when relying on manually generated labels as opposed to measured data. As GEOBERTje also manages to classify these cases correctly most of the time, this results in a slightly lower normalized accuracy score for some specific labels. Nonetheless, this outcome is a testament to the robust pattern-recognition capabilities of advanced language models even when trained on imperfect datasets.

While GEOBERTje sometimes omits certain lithology classes by assigning ``None'' for secondary or tertiary lithologies where other labels might be applicable, it rarely hallucinates. Conversely, hallucinations are a notable problem for the GPT-4 model results. More specifically, GPT-4 tends to invent lithology classes within the sand-group (e.g. \textit{``very calcareous whitish sand''} is classified by GPT-4 as \textit{``fine sand''}). This is a critical issue for downstream applications such as geological models: missing input data are generally far less problematic than the introduction of incorrect lithology classes which could lead to significant inaccuracies in the models.

\subsection{Accuracy versus training set size}
\begin{figure}[htbp]
	\centering
	\includegraphics[width=0.6\textwidth]{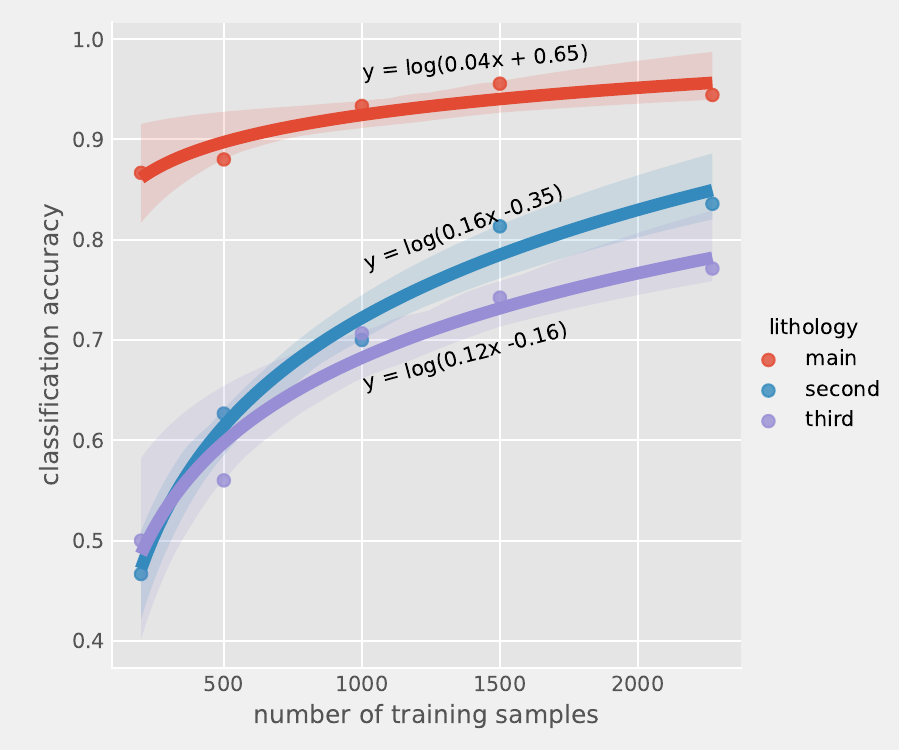}
	\caption{The impact of labeled training dataset size on GEOBERTje's lithology classification accuracy.}
	\label{fig:samles_annealing}
\end{figure}

We investigated the impact of the labeled training set size on classification accuracy because the availability of sufficient ground truth labels represents an important challenge for the potential of LLMs to act as an alternative to more traditional classification approaches. This provides insights in whether we possess adequate labeled data for the model finetuning and how many additional samples are required to attain specific performance levels.
Figure~\ref{fig:samles_annealing} represents the relation between the number of training 
data and the ability of the three models to correctly predict the main, second, and third lithologies. 
These results show that a relatively small number of training samples (e.g., $500$) suffice 
for GEOBERTje to achieve a reasonable accuracy of $0.90$ in predicting the main lithology. 
This suggests again that extensive geological context may not be necessary for main lithology prediction and elucidates why the zero-shot GPT-4 model yields a relatively high accuracy of $0.85$ in Figure~\ref{fig:accuracies}.

The accuracy of predicting the main lithology gradually increases by adding more training data 
before plateauing at around $1500$ training observations. 
However, this does not apply to the secondary lithologies as the accuracies continue to exhibit a significant improvement.
We found that increasing the training samples from $500$ to $2270$ yields a gain of approximately $20\%$ in accuracy. This underscores the significance of labeled data in attaining higher accuracy, especially for the second and third lithologies. Our results, based on the extrapolation of these logarithmic trends using $accuracy = \log size$ for the different lithologies, suggest an accuracy of $89.3\%$ would be achieved for $3000$ training observations for the second and $83.5\%$ at $3500$ training observations for the third lithology.

\subsection{Future work}
Future research can further enhance the model we developed. GEOBERTje's domain adaptation would improve further with training on a larger dataset. Additionally, we illustrated that the classifier would benefit from additional labeled data specifically focusing on phrases describing at least two different lithologies. Finally, linking the three classifiers to make them aware of each other's predictions could potentially improve the results.
Our modeling approach can also easily be transferred to different languages.
Moreover, the potential of GEOBERTje is broader than the lithological classification for which it is currently developed. The domain adaptation can be easily expanded by including a more diverse corpus of geological data (e.g. stratigraphical borehole interpretations or geological reports). This will result in improved embeddings that reflect geological context and semantic relations on a more holistic level. This would also open the door for a broader range of downstream applications such as lithostratigraphic mapping or resource estimation.

\section{Conclusions}\label{sec:conclusions}
Geological borehole descriptions, often collected over many decades by geological survey organizations, contain a wealth of information because they contain detailed textual descriptions of the subsurface composition. They are key input data for geological models and are used in several other fields (e.g., mineral exploration, groundwater management, geotechnical engineering, etc.). Yet, their unstructured text format poses a challenge when extracting all relevant features into a numeric format that is usable in computer models. 

This study takes advantage of the large language model (LLM) revolution, ushered in by the transformer architecture, to develop GEOBERTje: a domain adapted large language model trained on geological borehole descriptions from Flanders (Belgium) in the Dutch language. The domain adapted model extracts relevant information from borehole descriptions and represents it in a numeric vector space. Showcasing just one potential application of GEOBERTje, we finetune a classifier model on a limited number of manually labeled observations. The classifier categorizes borehole descriptions into a main, second and third lithology. We show that our classifier significantly outperforms both a rule-based regular expression script and GPT-4 of OpenAI. 

Adopting domain-specific LLMs in the field of geology, as exemplified by our study, underscores the transformative potential of these models in enhancing the accessibility of unstructured geological datasets. This innovation paves the way for integrating vast amounts of data, leading to more efficient and accurate geological analyses.

\section*{Acknowledgments}
We are grateful to Lorenz Hambsch for his research assistance on the GPT-4 prompt engineering and to Roel De Koninck and Katrien De Nil for helpful comments that improved the paper. The usual disclaimer applies.

This paper was partially supported by the project ``Technology Watch'' of the VLAKO reference task, ordered by the Bureau for Environment and Spatial development - Flanders, VPO. We are grateful to have had the opportunity to share the research within the open DOV network. 

\section*{Model Availability}
GEOBERTje is freely available on Hugging Face: ~\cite{geobertje2024}. 
\bibliography{biblio.bib}
\appendix
\begin{appendices}
	
\section{Other methods}\label{othermethods}
\subsection{Classification using rule-based scripting}\label{rulebased}
Up to now, geological borehole descriptions were translated into lithoclasses using a set of rule-based scripts \citep{vanharen2016, vanharen2023}. At the basis of these scripts is a dictionary providing the mapping of a large corpus of geological words and phrases on a lithoclass (e.g. 'flint pebbles' -> 'gravel'). Also, this dictionary indicates whether a certain phrase represents a main, or rather a secondary lithology (e.g. 'clayey' -> secondary lithology). 
Additionally, a series of regular expressions are implemented to handle more complex linguistic structures, such as phrases where the adverb defining the grain size of the sand class is not positioned directly adjacent to the word sand (e.g. 'fine, greenish sand' -> 'fine sand). Finally, logic was incorporated to accurately classify descriptions that specifically denote the absence of a lithology (e.g. 'sand without gravel'). This rule-based framework harnesses a lot of domain specific-knowledge to achieve an automated classification of lithological descriptions, but it is limited by its inability to fully capture the diverse and intricate linguistic nuances present in many geological descriptions.

\subsection{Classification using ChatGPT}\label{gpt}
In addition to comparing our domain adapted and finetuned GEOBERTje classifier to the rule-based script, we further evaluate it by contrasting its outputs with those produced by the popular GPT model. The tests were performed on version gpt-4-0125-preview, with the temperature set to 0. Our objective was to determine whether the intricate process of domain adaptation and finetuning could potentially be circumvented by harnessing the capabilities of this prevalent commercial model. Should GPT's performance parallel that of GEOBERTje, it would suggest that complex geological language processing can be executed by simple prompt engineering. This prospect opens up the possibility for geologists without deep technical expertise to execute advanced tasks, diminishing the reliance on specialized data scientists. Therefore, this test was conducted by using our labeled validation set to engineer a prompt aimed at achieving optimal classification performance. The complete final prompt is provided in appendix \ref{appendix: prompt}. The performance of ChatGPT was subsequently assessed by comparing its classification results with those of the GEOBERTje classifier on the same test set. Although finetuning the GPT model is currently possible, this technique was not employed as it would contradict our objective of exploring the potential of a non-technical solution for this kind of task.
	
\subsubsection{GPT-4 prompt}\label{appendix: prompt}
Below is the GPT-4 prompt, employed using version gpt-4-0125-preview, with the temperature set to zero:
\begin{quote}
Classify geological drill core sample descriptions into these material types:
\begin{itemize}
    \item \texttt{fijn\_zand, grof\_zand, middel\_zand, zand\_onb, silt, grind, veen, klei, leem, none}.
    \item \texttt{grind} includes all things related to stones or pebbles.
    \item Use \texttt{zand\_onb} for any sand, except if it is clearly defined as fine, middle, or coarse sand. Color does not matter.
    \item If the main substrate is sandy then \texttt{zand\_onb} should be the second class.
    \item There can only be one type referencing the sand per row.
    \item ``Hetzelfde'' or ``idem'' means use the same classification as the row before this one.
    \item Words ending on ``achtige'' always refers to the second or third class.
    \item Make 3 classifications. The first column has the type of material that is the main ingredient in this sample. The second column has the next most common type, and the third column has the least common type.
    \item Respond only with this list of classifications. Use ``none'' if a material is not clearly indicated.
    \item Only use one of the types or none, nothing else.
    \item Always have 3 output types. Multiple nones are allowed.
    \item The input list has each row between triple quotes.
\end{itemize}
\end{quote}

\section{Confusion matrices}\label{appendix: confusion matrix}
The figure below provides a 
detailed breakdown of the model performance across different classes. 
Each confusion matrix sets out the proportion of predicted observations 
versus the actual observations for each main, second, third lithology classes. 
Notably, the row values in each matrix are normalized by the true labels, with the diagonal numbers reflecting the percentage of correctly predicted instances per class, 
thereby revealing class-specific performance strengths and weaknesses.

\begin{figure}[htbp]
	\centering
	\includegraphics[width=1.0\textwidth]{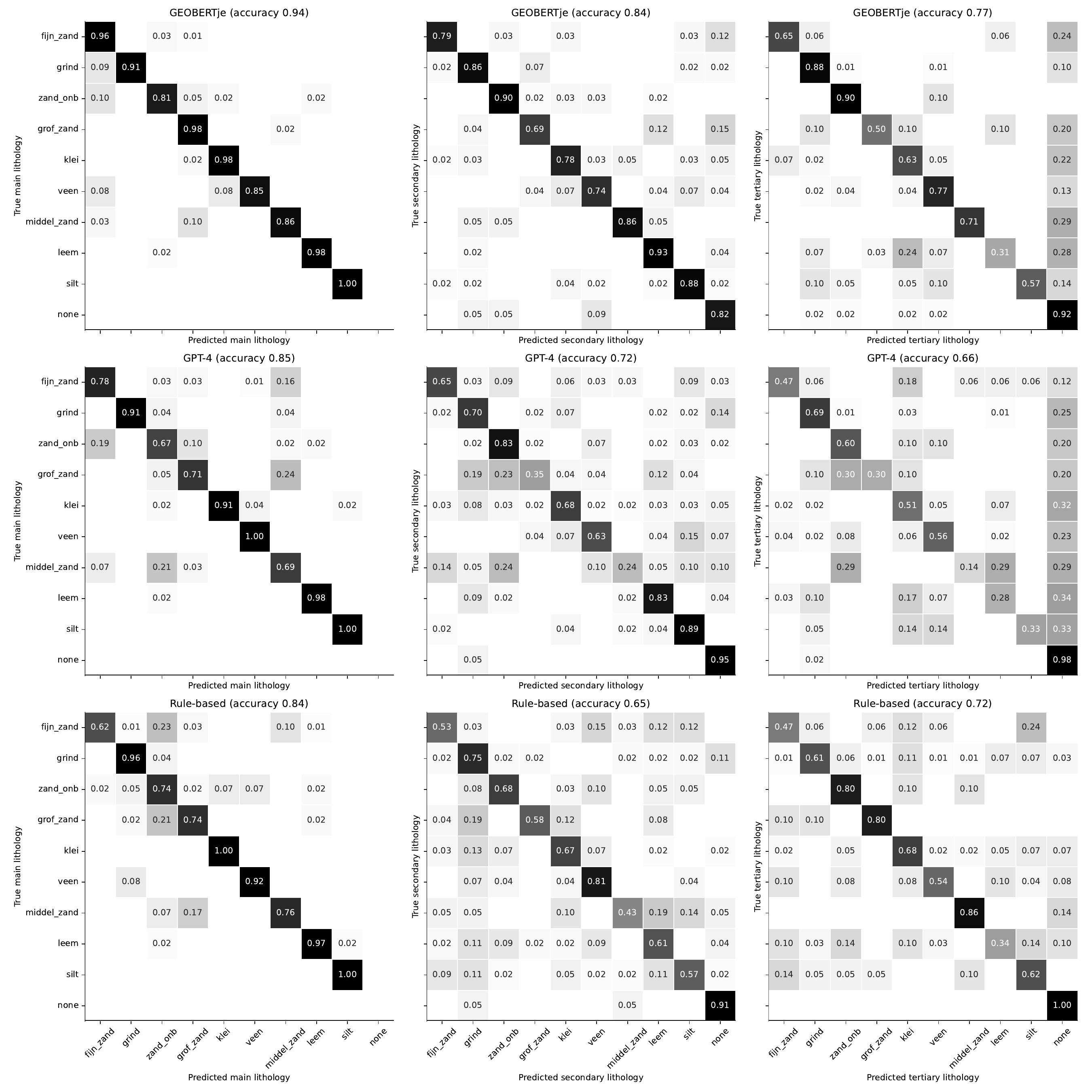}
	\caption{Confusion matrices.}
	\label{fig:confusion_matrices}
\end{figure}

\end{appendices}
\end{document}